\documentclass[letterpaper, 10 pt, conference]{ieeeconf}  

\IEEEoverridecommandlockouts                              

\usepackage{graphicx}
\usepackage{amsmath}
\usepackage{amssymb}
\usepackage{comment}
\usepackage{amsfonts}
\usepackage{booktabs,multirow}
\usepackage{bm}
\usepackage{amsmath,amssymb,color}
\usepackage{xcolor}
\usepackage{hyperref}

\newcommand{\gao}[1]{{\textcolor{black}{#1}}}

\title{\LARGE \bf
Accurate Grid Keypoint Learning for Efficient Video Prediction 
}

\author{Xiaojie Gao$^{*}$, Yueming Jin$^{*}$, Qi Dou, Chi-Wing Fu, and Pheng-Ann Heng
\thanks{* Equal contribution.}
\thanks{This work was supported by Hong Kong Research Grants Council with Project No. CUHK 14201620.}
\thanks{X. Gao, Y. Jin, Q. Dou, C.-W. Fu, and P.-A. Heng are with the Department of Computer Science and Engineering, The Chinese University of Hong Kong, China. Q. Dou is also with the CUHK T Stone Robotics Institute.
P.-A. Heng is also with Guangdong-Hong Kong-Macao Joint Laboratory of Human-Machine Intelligence-Synergy Systems, Shenzhen Institutes of Advanced Technology, Chinese Academy of Sciences, China.
\emph{Corresponding author at: qdou@cse.cuhk.edu.hk (Qi Dou).}}%
}

\begin{document}

\maketitle
\thispagestyle{empty}
\pagestyle{empty}

\begin{abstract}

Video prediction methods generally consume substantial computing resources in training and deployment, among which keypoint-based approaches show promising improvement in efficiency by simplifying dense image prediction to light keypoint prediction. 
However, keypoint locations are often modeled only as continuous coordinates, so noise from semantically insignificant deviations in videos easily disrupt learning stability, leading to inaccurate keypoint modeling.
In this paper, we design a new grid keypoint learning framework, aiming at a robust and explainable intermediate keypoint representation for long-term efficient video prediction. 
We have two major technical contributions. 
First, we detect keypoints by jumping among candidate locations in our raised grid space and formulate a condensation loss to encourage meaningful keypoints with strong representative capability. 
Second, we introduce a 2D binary map to represent the detected grid keypoints and then suggest propagating keypoint locations with stochasticity by selecting entries in the discrete grid space, thus preserving the spatial structure of keypoints in the long-term horizon for better future frame generation.
Extensive experiments verify that our method outperforms the state-of-the-art stochastic video prediction methods while saves more than 98\% of computing resources. 
We also demonstrate our method on a robotic-assisted surgery dataset with promising results. 
Our code is available at \url{https://github.com/xjgaocs/Grid-Keypoint-Learning}.
\end{abstract}

\section{Introduction}

Unsupervised video prediction aims to synthesize future frames based on observations in previous frames without requiring any annotation~\cite{byeon2018contextvp,kumar2020videoflow}.
Its look-ahead capability enables essential board applications in robotic navigation, video surveillance, and autonomous vehicles~\cite{finn2017deep,jin2018varnet}. 
Through timely anticipation of the future, it aids intelligent decision making and also emergency-response system~\cite{gao2020automatic}.
Significantly, precisely predicting videos for a more extended period while upholding computing efficiency can further widen the applicability of deployments on mobile robots and domestic service robots.
However, generating future frames with plausible motion dynamics is very challenging due to the difficulty of processing the high-dimensional video data~\cite{villegas2017learning}.
Thus, predictions by existing approaches tend to miss critical visual details and suffer from motion blurry and image distortion~\cite{jin2020exploring}. 
These issues are even amplified with increases in prediction steps.

\begin{figure}[t]
	\includegraphics[width=0.45\textwidth]{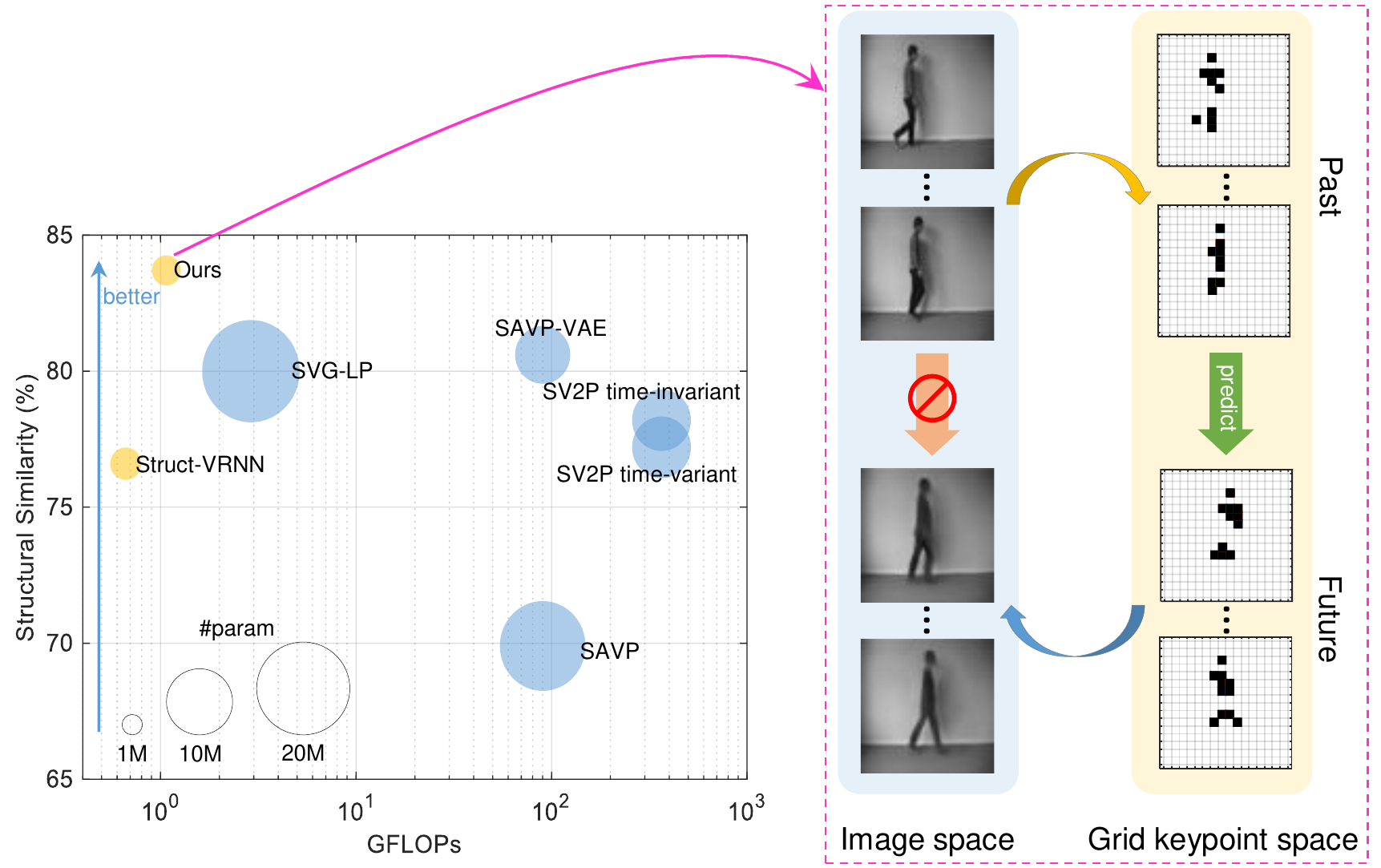}
	\centering
	\caption{By transforming frames into our smartly designed grid keypoint space, accurate keypoint configurations can be predicted using our framework, thereby enabling the best video prediction performance efficiently.
	}
	\label{fig:overview}
	\vspace{-4mm}
\end{figure}


To capture various tendencies in the future, stochastic video prediction approaches were developed by defining a prior distribution over a set of latent variables, allowing different ways of sampling the distribution.
Most of them focused on the direct pixel-wise synthesis of predicted future frames, known as the image-based prediction model.
For this stream of image-based prediction models, recent studies pointed out that increasing the model scale could improve performances~\cite{castrejon2019improved,villegas2019high}.
However, a vast prediction model, on the other hand, would consume extremely large memory and energy, being impractical for real-world deployments.

One promising direction to enhance efficiency is by reducing the prediction space from dense pixel-wise images to some high-level representations, such as {\em keypoint coordinates\/}~\cite{jakab2018unsupervised,zhang2018unsupervised}, where future frames are synthesized by analogy making with the predicted high-level representations and a reference frame.
This representation creates an explicit high-level structure and simplifies the dynamics to be learned, thereby substantially decreasing the model complexity. 
As shown in Fig.~\ref{fig:overview}, keypoint-based methods (denoted in yellow) require much fewer parameters and computing resources than image-based approaches (denoted in blue).
However, there is still a performance gap between the state-of-the-art keypoint-based model, i.e., Struct-VRNN~\cite{minderer2019unsupervised}, and image-based methods.


The inferior results of existing keypoint-based video prediction models are due to two significant problems.
First, keypoints are detected and represented in a continuous coordinate space,
where spatial relationships and constraints transferred from video frames could hardly be preserved without sophisticated regularization. 
Hence, the keypoints exhibit limited representative capacity, and artifacts are produced in synthesized frames when transforming information back to image space.
Second, they propagate keypoints in temporal dimensions by regressing continuous coordinates, thus further destroy the keypoint structures due to inaccurate predictions.
Notably, for long-term predictions, the adverse effect becomes more severe given that the compounding of errors accumulates over time.

To address these critical issues mentioned above, we propose a novel grid keypoint representation learning framework for long-term video prediction with various possibilities by enhancing the keypoint representation capacity and coordinate propagation reliability. 
Our main contributions are: 
\textbf{1).} 
To regularize the detected keypoints, we develop a new gridding operation to compress the keypoint coordinates from infinite and continuous space to finite and discrete grid space, as shown in Fig.~\ref{fig:overview}. 
To our best knowledge, this is the first method that conducts grid keypoint learning for video prediction. 
\textbf{2).} We propose a novel condensation loss to encourage the model to concentrate on the most informative region. 
Combining with the gridding operation, it vastly promotes the representative capability of keypoints, thus concentrated and meaningful keypoints are inferred. 
\textbf{3).} To facilitate keypoint coordinate propagation, we devise a 2D binary map to represent the spatial relationships of keypoints and predict future keypoint by choosing its location in the finite grid space, transferring the prediction task from previous regression to classification.
Thus, the compounding of coordinate errors are substantially reduced to enable future frame generation with high-fidelity. 
\textbf{4).} Extensive results demonstrate that our method maintains keypoint structures in long-term horizons and achieves superior performances and efficiency over the state-of-the-art stochastic video prediction models. 
We also illustrate the great potential of our method on robotic-assisted surgery.  

\begin{figure*}[t]
	\includegraphics[width=\textwidth]{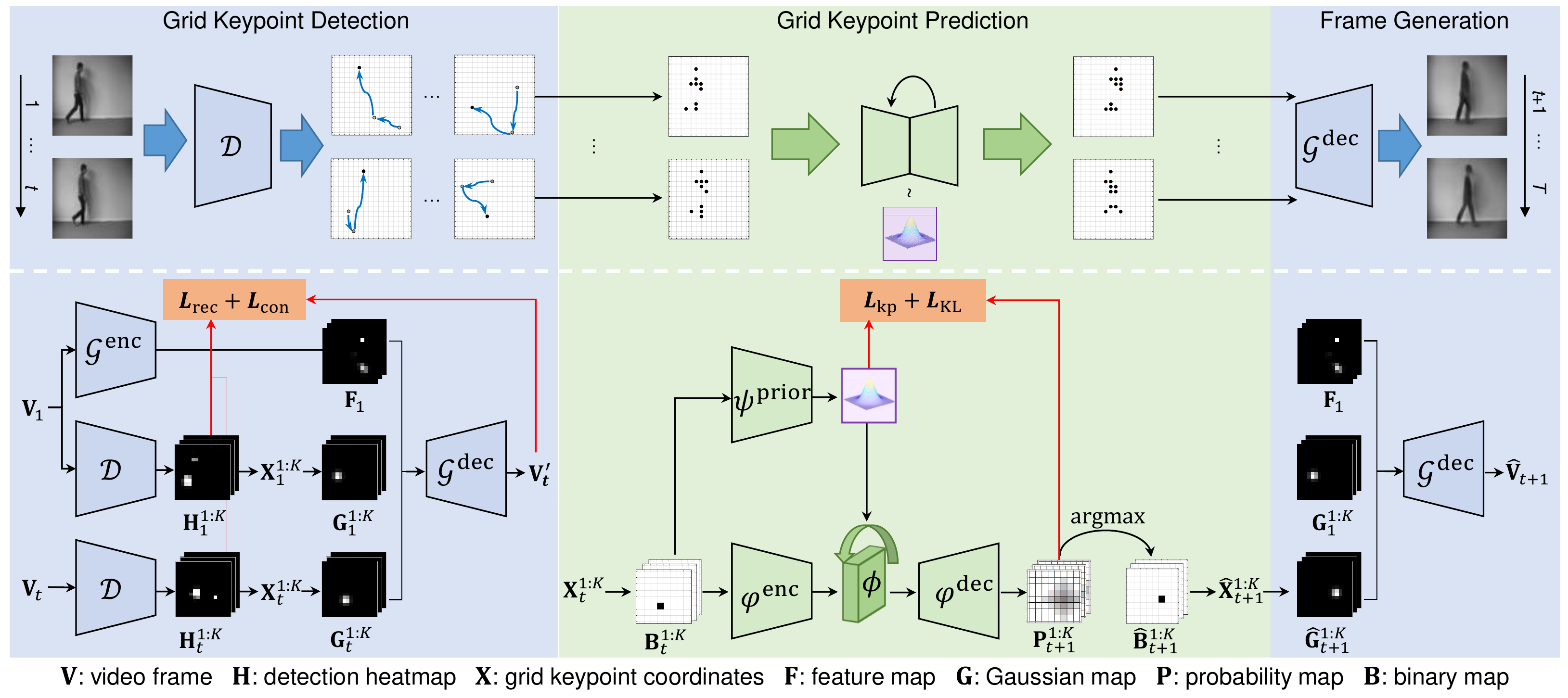}
	\centering
	\caption{Overview of our grid keypoint learning framework. 
	Our pipeline contains three stages: grid keypoints of observed frames are first detected in the canonical grid space;
	future keypoint coordinates are propagated by choosing the grid locations with maximum probabilities;
	future frames are generated by translating the predicted keypoints via analogy making.
	}
	\label{fig:framework}
	\vspace{-5mm}
\end{figure*}

\section{Related work}

Existing video prediction methods can be divided into two categories: deterministic and stochastic prediction.
In this study, we focus on the latter one that could learn to represent a diverse future.
Kalchbrenner et al.~\cite{kalchbrenner2017video} presented an autoregressive model that directly maximizes the log-likelihood of the data at a slow speed.
Kumar et al.~\cite{kumar2020videoflow} proposed
a flow-based method to allow direct optimization of the data likelihood, which might fail to capture complex motion. 
GAN-based models were also applied to model inexplicit data distribution.
Tulyakov et al.~\cite{tulyakov2018mocogan} used GANs for unconditional video generation, however, using adversarial losses generally encounters training difficulties such as mode collapse.
Other vital foundations of probabilistic models are VAE and variational recurrent neural network (VRNN)~\cite{chung2015recurrent}.
Babaeizadeh et al.~\cite{babaeizadeh2018stochastic} applied VAE on video prediction by encoding the entire video sequence to estimate a posterior distribution.
A stochastic video generation model using learned prior (SVG-LP) instead of the standard Gaussian prior was proposed in~\cite{denton2018stochastic}.  
Lee et al.~\cite{lee2018stochastic} combined GAN with VAE to produce sharp and realistic future frames.
Remarkable performance boosts were achieved by increasing the expressive capacity of the latent variables~\cite{castrejon2019improved,villegas2019high}, however, the resulted models were too big to be trained with general computers. 
The above methods generally rely on image-autoregressive processes for updating temporal recurrence and suffer from gradually noisy outputs as time step increases.
Franceschi et al.~\cite{franceschi2020stochastic} proposed a computationally appealing method by separating the temporal dynamics from frame synthesis inexplicitly.
Disentangling hidden dynamics and appearance representation explicitly, keypoint-based video prediction methods were suggested~\cite{minderer2019unsupervised,kim2019unsupervised}, which first represent images with keypoints in an unsupervised manner and then synthesize future frames given predicted keypoints.
 
Unsupervised keypoint learning was first proposed in images~\cite{jakab2018unsupervised,zhang2018unsupervised}, where a representational bottleneck forces a neural network to encode structural information into several keypoints with continuous coordinates.
To predict the dynamics of detected keypoint sequences for generating future videos, coordinates are regressed using a basic VRNN architecture~\cite{minderer2019unsupervised} or a stochastic sequence-to-sequence model conditioning on class labels~\cite{kim2019unsupervised}.
Villegas et al.~\cite{villegas2017learning} also predicted keypoint coordinates with a sequence-to-sequence model based on LSTM yet in a deterministic way, which gained good outcomes thanks to the manually annotated keypoints. 
These approaches employ recurrent architectures to regress the coordinates represented by 1D vectors, producing unsatisfying results due to inaccurate predictions of keypoint coordinates. 
\gao{Since the keypoints generated in an unsupervised manner could not maintain the point correspondence, such as confusion about left and right legs of humans, these keypoints are more inclined to suffer from propagation noise, thereby leading to weird results.}
How to more accurately predict future keypoints without human annotations is of great importance to produce more realistic videos.

\section{Method}
Fig.~\ref{fig:framework} illustrates an overview of our proposed grid keypoint learning framework. 
Given observed video frames $\mathbf{V}_{1:t}$, we first detect corresponding keypoints in the proposed grid space, followed by our grid keypoint prediction network for accurate coordinate propagation. 
By elegantly designing binary maps, our method substantially decreases the accumulated errors of keypoint coordinates, thus generates future frames $\mathbf{\hat V}_{t+1:T}$ with high-fidelity.


\subsection{Keypoint Detection in Grid Space}
\label{sec:detnet}
Given a video frame $\mathbf{V}_{t}\in \mathbb{R}^{C\times H\times W}$, we aim to represent it with $K$ keypoints without supervision, which try to restore original image details as many as possible helped by a decoder network. 
Previous keypoint detection methods employed a bottleneck to reconstruct the frame $\mathbf{V}_{t}$ based a reference frame $\mathbf{V}_1$ by analogy making using corresponding keypoints~\cite{jakab2018unsupervised,minderer2019unsupervised}. 
Instead of detecting keypoints with continuous coordinates, we propose to identify appropriate keypoints in a novel grid plane because image pixels are stored in standard grid forms, and there is no need to produce keypoints with higher resolution than images. 
Moreover, our grid constraint serves as regularization by sparing minimum distances among keypoints to prevent overfitting, which promotes the generalization ability of our keypoints to represent unseen pictures. 
However, searching grid keypoints discretely is intractable due to exponential complexity.

To meet this challenge, we devise a novel grid keypoint learning, which updates keypoint locations in a grid space, denoted as $\mathcal{I}_{HW}$ with a resolution of $H\times W$.
With $\mathbf{V}_{t}$ as input, our keypoint detector $\mathcal{D}$ tries to output $K$ grid keypoint coordinates $\mathbf{X}_{t}^{1:K}$.
As shown in Fig.~\ref{fig:framework}, $\mathcal{D}$ first produces $K$ heatmaps $\mathbf{H}_t^{1:K}$ activated by a sigmoid function, which are transformed into intermediate keypoint coordinates $\mathbf{\bar X}_t^{1:K}=[\bar x_t^{1:K}, \bar y_t^{1:K}]$ by computing the spatial expectations of the heatmaps.
The $K$ keypoints with continuous coordinates are then pushed to their nearest grid points respectively to generate grid keypoints $\mathbf{X}_t^{1:K}=[x_t^{1:K}, y_t^{1:K}]$. 
A trivial way for this operation is using round operation, but gradients cannot be backpropagated through the network for parameter update. 
Instead, to enable training of $\mathcal{D}$, we realize this pushing operation by elegantly adding the $k$-th intermediate keypoint $\mathbf{\bar X}_t^{k}$ with a constant difference:
\begin{equation}
     \Delta\mathbf{X}_t^{k} = \arg\min_{\mathbf{\widetilde X}}||\mathbf{\widetilde X}-\mathbf{\bar X}_t^{k}||^2_2-\mathbf{\bar X}_t^{k} ,
\end{equation}
\gao{where $\mathbf{\widetilde X}$ is the coordinate of a grid point in $\mathcal{I}_{HW}$.}
Then, $\mathbf{X}_t^{1:K}$ are 
represented with Gaussian-shaped blobs at their grid locations to form Gaussian maps $\mathbf{G}_t^{1:K}$. 
To bring the semantic information for reconstruction, we concatenate $\mathbf{G}_t^{1:K}$ with the appearance feature maps of the reference frame $\mathbf{F}_1$ output from an encoder network $\mathcal{G}^\mathrm{enc}$.
Gaussian map of the reference frame $\mathbf{G}_1^{1:K}$ is also concatenated for inpainting the background regions.
The final results are input to a decoder network $\mathcal{G}^\mathrm{dec}$ to reconstruct ${\mathbf V}_t$ by generating ${\mathbf V}_t'$ to finish the forward pass.
As for the backward pass to update network parameters, $\mathcal D$ and $\mathcal G:\{\mathcal{G}^\mathrm{enc}, \mathcal{G}^\mathrm{dec}\}$, 
are jointly training to optimize an $\ell_2$ reconstruction loss:
\begin{equation}
     \mathcal{L}_\mathrm{rec}=\sum_{t=1}^{T}\|\mathbf{V}_t-{\mathbf V}_t'\|_2^2.
\end{equation}
\gao{Note that $\mathbf{F}_t$ could also be used as a reference frame, and slightly better results could be obtained.}



To this end, the keypoint detector $\mathcal D$ and the image synthesizer $\mathcal G$ constitute an autoencoder architecture to encode frame $\mathbf{V}_t$ into keypoint-based representations.
The gradients from $\mathcal{L}_\mathrm{rec}$ encourage $\mathcal D$ to adjust its parameters to generate optimal keypoint patterns in the grid space. 
We demonstrate in Section~\ref{sec:ablation_rec} that our grid keypoints exhibit a more robust capability to restore original images than keypoints with continuous coordinates by helping preserve a lot more details of the original frame $\mathbf{V}_t$.

\noindent\textbf{Condensation Loss for Robust Grid Keypoint.} 
To interpret a single keypoint, the interesting areas in each heatmap should be concentrated, and the activation values of the irrelevant regions are relatively low.
As each heatmap $\mathbf H_t^k$ is activated by a sigmoid function, the optimal structure of each heatmap shall contain a single entry as value 1 and the rest as 0, showing the maximum contrast.
By contrast, a heatmap with the same values generates the most ambiguous detection (the worst condition), where $\max(\mathbf{H}_t^k)$ is equal to $\mathrm{mean}(\mathbf{H}_t^k)$.
To make the keypoints sparse and robust to noisy images, we introduce a new regularization term called condensation loss.
It is devised by broadening the gap between $\max(\mathbf{H}_t^k)$ and $\mathrm{mean}(\mathbf{H}_t^k)$ for all produced heatmaps to enhance centralized distributions of heatmaps:
\begin{equation}
\mathcal{L}_\mathrm{con}=-\sum_{t}\sum_k(\max(\mathbf{H}_t^k)-\mathrm{mean}(\mathbf{H}_t^k)).
\end{equation}
In practice, we find that only optimizing the worst heatmap among the $K$ channels for all time steps also creates an excellent performance. 
In Section~\ref{sec:ablation_rec}, we show that $\mathcal{L}_\mathrm{con}$ contributes to a better reconstruction performance by facilitating concentrated keypoint configurations.


Thus, our grid keypoint detection network is trained by jointly optimizing $\mathcal{D}$ and $\mathcal{G}$ using a combined loss:
\begin{equation}
 \mathcal{L}_\mathrm{det}=\mathcal{L}_\mathrm{rec}+\lambda \mathcal{L}_\mathrm{con},
\end{equation}
where $\lambda$ is a constant to balance the two terms. 
Note that the well-trained $\mathcal{G}^\mathrm{dec}$ is directly reused in the generation of future frames.

\subsection{Grid Keypoint Prediction via Binary Map} 

With our detected grid keypoints, we develop a keypoint prediction method to alleviate side effects from compounding of keypoint coordinate errors for realistic future frame generation.  
Previous methods predict future keypoints by regressing the coordinates in 1D vector form~\cite{villegas2017learning, minderer2019unsupervised,kim2019unsupervised}, which can hardly maintain spatial structures of keypoints. 
Although 2D Gaussian maps could express spatial information of keypoints, they still suffer from the accumulation of errors severely due to regressing the continuous coordinates. 
We propose to select keypoint locations in the finite grid space, which eschews accumulated errors due to continuous coordinate regression.

We first devise a novel binary map for precise representations of keypoints in the grid space.
Concretely, given a detected keypoint with coordinate as $[x_t^k, y_t^k]$, we scale it to find its corresponding entry in an $H\times W$ grid map and make the entry be 1 while the rest 0, forming our binary map $\mathbf B_t^k\in \{0,1\}^{H\times W}$ to represent the $k$-th keypoint of $\mathbf X_t$. 
As a kind of sparse representation, our binary map shares a similar spirit with AlphaGo~\cite{silver2017mastering} that represents a stone on the board of Go. 
As shown in Fig.~\ref{fig:framework}, we indicate each keypoint location of $\mathbf X_t$ by a single channel of $\mathbf{B}_t$, which further inputs to our keypoint prediction network.

To reduce the coordinate error in prediction, we propose to choose keypoint locations indicating their discrete coordinates in the finite grid space rather than regress continuous coordinates. 
Therefore, we formulate the keypoint coordinate prediction as a classification task. 
As shown in Fig.~\ref{fig:framework}, our keypoint prediction network takes an input as the binary maps $\mathbf B_t^{1:K}$ and outputs probability maps $\mathbf P_{t+1}^{1:K}$ to specify the presence of all keypoints over possible positions for the next time step. 
An argmax operation is used to determine the predicted keypoint coordinates $\mathbf{\hat X}_{t+1}^{1:K}$ and binary maps $\mathbf{\hat B}_{t+1}^{1:K}$ by selecting entries with the maximum probabilities in $\mathbf P_{t+1}^{1:K}$ for each keypoint.
The binary maps $\mathbf{\hat B}_{t+1}^{1:K}$ are also taken as the input to our prediction model when $\mathbf B_{t+1}^{1:K}$ are not available during testing.
Compared to coordinate regression methods, our suggested style can purify a large amount of noise existing in raw outputs of the network by forming standard inputs as binary maps. 
To train our model, we leverage the cross-entropy loss to measure the prediction error between the predicted $\mathbf {P}_{t+1}$ and ground truth binary maps $\mathbf B_{t+1}$, which is derived from our well-trained grid keypoint detection network. We define the loss as
\begin{equation}
	\mathcal{L}_\mathrm{kp}= -\sum^{T-1}_{t=1}\mathbf{B}_{t+1}\log\mathbf P_{t+1}.
\end{equation}

\begin{figure*}[t]
	\includegraphics[width=0.9\textwidth]{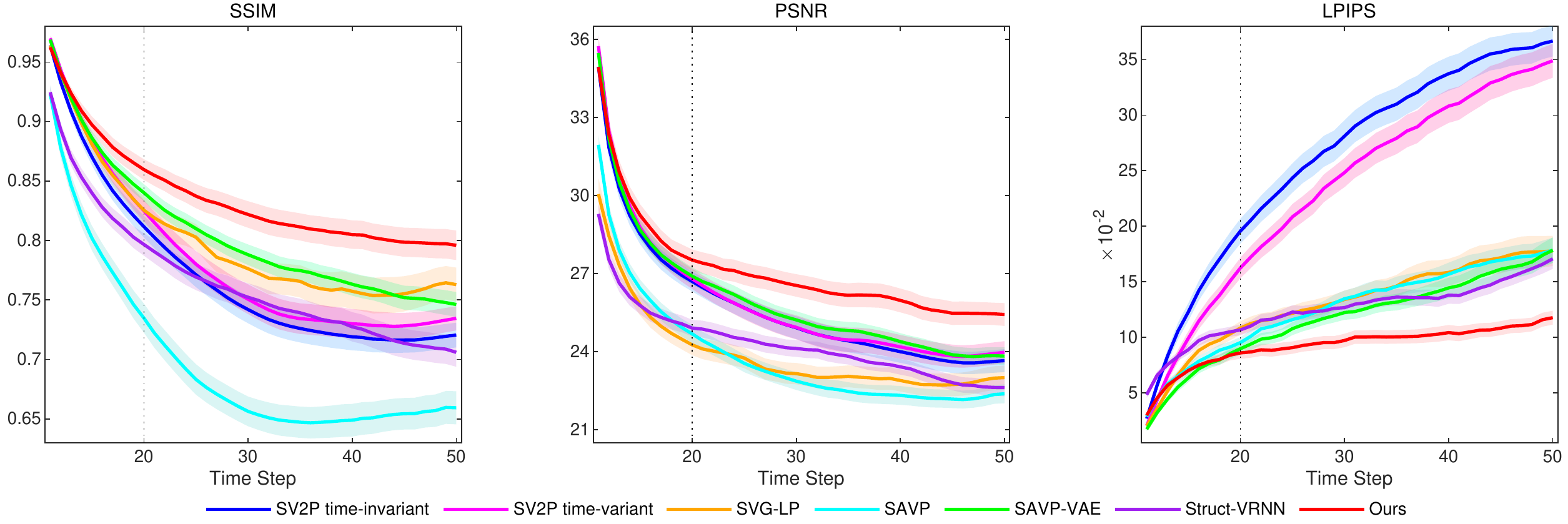}
	\centering
	\caption{Quantitative evaluation with respect to each time step for all models on the KTH dataset. 
		The	models are conditioned on the first 10 frames and predict the following 40 frames.
		The vertical dotted line indicates the time step the models were trained to predict up to. 
		Mean SSIM, PSNR, and LPIPS over all test videos are
		plotted with 95\% confidence interval shaded. 
		Higher SSIM, PSNR and lower LPIPS indicate better performances.
}
	\label{fig:kth_perframe}
	\vspace{-5mm}
\end{figure*}

With the above formulated coordinate prediction scheme, we extend it to consider the dynamics of keypoints and account for stochasticity in the future.
We establish our stochastic keypoint prediction network based on VRNN architecture~\cite{chung2015recurrent}.
The core insight is referring to a latent belief $\mathbf{z}$ to predict possible keypoint locations, where the latent belief $\mathbf{z}\in \mathbb R^{H/4\times W/4}$ is a single-channel response map~\cite{babaeizadeh2018stochastic} to model the stochasticity in keypoint sequences.
It is conditioned on the information of all previous frames recorded by hidden states of an RNN.
To model the spatio-temporal relations of keypoints on binary maps, we employ a convolutional LSTM (convLSTM)~\cite{shi2015convolutional} denoted as $\phi$ to generate hidden states $\mathbf h_t\in\mathbb R^{64\times H/4\times W/4}$.
In the prediction of keypoint at time step $t+1$, the prior latent belief $\mathbf{z}_{t+1}$ observes the information from $\mathbf{B}_1$ to $\mathbf{B}_t$ modeled by $\mathbf h_t$:
\begin{equation}
	p(\mathbf{z}_{t+1}|\mathbf{B}_{1:t},\mathbf{z}_{1:t})=\psi^\mathrm{prior}(\mathbf{h}_{t}).
\end{equation}
The posterior belief of $\mathbf z_{t+1}$ is obtained given additional information of time step $t+1$:
\begin{equation}
	q(\mathbf{z}_{t+1}|\mathbf{B}_{1:t+1},\mathbf{z}_{1:t})=\psi^\mathrm{post}(\mathbf{B}_{t+1}, \mathbf{h}_{t}),
\end{equation} 
where $\psi^\mathrm{prior}$ and $\psi^\mathrm{post}$ are our prior and posterior networks, respectively, to output the expectation and standard deviation of Gaussian distributions. 
With the latent belief $\mathbf{z}_{t+1}$, a keypoint decoder $\varphi^\mathrm{dec}$ predicts the keypoints of the next time step by
\begin{equation}
	p(\mathbf{B}_{t+1}|\mathbf{z}_{1:t+1},\mathbf{B}_{1:t})=\varphi^\mathrm{dec}(\mathbf{z}_{t+1}, \mathbf{h}_{t}).
\end{equation}
Finally, the hidden states are updated by incorporating newly available information to close the recurrent loop:
\begin{equation}
\mathbf{h}_{t+1}=\phi\left(\varphi^\mathrm{enc}(\mathbf{B}_{t+1}), \mathbf{z}_{t+1}, \mathbf{h}_{t}\right),
\end{equation}
where $\varphi^\mathrm{enc}$ is an encoder for size reduction.
During training, the recurrence in $\phi$ is updated using $\mathbf{B}_{1:T}$ and the posterior belief output by $\psi^\mathrm{post}$. 
When $\mathbf{B}_{t+1:T}$ is no more available during the inference stage, the predicted binary maps $\mathbf{\hat B}_{t+1:T}$ are applied with the prior belief from $\psi^\mathrm{prior}$ that is fitted to $\psi^\mathrm{post}$ during training.
Our VRNN architecture is optimized by maximizing the evidence lower bound (ELBO) using the re-parametrization trick~\cite{kingma2013auto}:
\begin{equation}
    \label{eqn:elbo}
	\sum_{t=1}^{T-1}\mathbb{E}[\log p(\mathbf{B}_{t+1}|\mathbf{z}_{1:t+1},\mathbf{B}_{1:t})-\beta\mathrm{KL}(q(\mathbf z_{t+1})||p(\mathbf z_{t+1}))],
\end{equation}
where $\beta$ is used to keep a balance between the reconstruction and prior fitting errors. 

In our keypoint prediction network, we replace the reconstruction term in ELBO by our keypoint prediction loss $\mathcal{L}_\mathrm{kp}$, and the overall training loss is given by
\begin{equation}
\mathcal{L}_\mathrm{pred}=\mathcal{L}_\mathrm{kp}+\beta\mathcal{L}_\mathrm{KL},
\end{equation}
where $\mathcal{L}_\mathrm{KL}=\sum_{t=1}^{T-1}\mathrm{KL}(q(\mathbf z_{t+1})||p(\mathbf z_{t+1}))$ is the KL-divergence between prior and posterior probabilities. 
Finally, the predicted $\mathbf {\hat B}_{t+1}$ with stochasticity is input to $\mathcal G^\mathrm{dec}$ for diverse future frame generation.
Our smart design yields substantially less noise in keypoint coordinate propagation and the synthesized future frames enjoy high fidelity, which is verified in Section~\ref{sec:ablation_pred}.

\section{Results}

\subsection{Experiment Setup}
\noindent\textbf{Datasets.}
We extensively validate our method on two datasets commonly used for the evaluation of stochastic video predictions. 
The KTH dataset~\cite{schuldt2004recognizing} contains real-world videos of 25 people performing six types of actions, and we use persons 1-16 for training and 17-25 for testing.
We use the same setting as~\cite{denton2018stochastic} to predict the subsequent 10 frames based on 10 observed frames. 
The prediction range extends to 40 frames in testing. 
The Human3.6M dataset~\cite{ionescu2013human3} also contains video sequences of human actors performing different actions.
We split the training and testing set and follow the experimental settings in~\cite{minderer2019unsupervised}.
During the training, models are conditioned on 8 observed frames and predict 8 frames. 
When testing, models predict 42 frames.

\noindent\textbf{Metrics.} 
For quantitative evaluation, we employ three commonly-used frame-wise metrics and average over time:
Structural Similarity (SSIM)~\cite{wang2004image}, Peak Signal-to-Noise Ratio (PSNR), and Learned Perceptual Image Patch Similarity (LPIPS)~\cite{zhang2018unreasonable}. 
Unlike SSIM and PSNR, LPIPS is a perceptual metric in the feature level of convolutional neural networks, which is more relevant to human judgment.
For SSIM and PSNR, higher values indicate better results, while lower results are preferred for LPIPS.
We also adopt Fr\'echet Video Distance (FVD)~\cite{unterthiner2018towards} to evaluate the results in video-level. 

\noindent\textbf{Implementation Details.} 
In all datasets, the keypoint grid resolution is set to $64\times 64$, and the size of the hidden state map is $16\times 16$.
We train our models using the Adam optimizer~\cite{kingma2014adam} with an initial learning rate of 1e-3 and an exponential decay rate of 0.25. 
We empirically set the keypoint number as $K=12$ (see Section~\ref{sec:ablation_num} for ablation study). 
We set $\lambda $ and $\beta$ to 0.01 and 0.1, respectively.

\subsection{Comparison with Existing Methods}
We compared our model with several state-of-the-art image-based stochastic video prediction approaches using image-autoregressive recurrent networks, including two variants of SV2P~\cite{babaeizadeh2018stochastic}, SVG-LP~\cite{denton2018stochastic}, SAVP, and its VAE-only variant~\cite{lee2018stochastic}. 
Additionally, we compare with the latest keypoint-based video prediction method Struct-VRNN~\cite{minderer2019unsupervised}.
For methods~\cite{babaeizadeh2018stochastic,denton2018stochastic,lee2018stochastic}, we obtain the results by directly running the available pre-trained models that authors released online.
For Struct-VRNN~\cite{minderer2019unsupervised}, we reimplement the method based on their released code under the same experimental settings.
Our evaluation process also strictly follows the previous methods~\cite{babaeizadeh2018stochastic,denton2018stochastic,lee2018stochastic}, where we first perform 100 random samples for each test sequence and choose the best scores with respect to the ground truth for each metric.
Average values over the entire test set are reported as the final results. 
Notably, we make the test sequences of all models precisely the same for a fair comparison.

\begin{table}[t]
	\centering
	\caption{Average comparison results of different methods on the KTH dataset. 
		The best results are marked in bold.
	}
	\label{tab:KTH}
	\resizebox{0.45\textwidth}{!}{
		\begin{tabular}{l|ccccc}
			\toprule
			 Method 
			& SSIM$\uparrow$ & PSNR$\uparrow$ & LPIPS$\downarrow$ & FVD$\downarrow$ & \#param\\ 
			\midrule
			
			SV2P time-invariant~\cite{babaeizadeh2018stochastic} & 0.772 & 25.70 & 0.260 & 253.5 & 8.3M\\
			SV2P time-variant~\cite{babaeizadeh2018stochastic} & 0.782 & 25.87 & 0.232 & 209.5 & 8.3M\\
			SVG-LP~\cite{denton2018stochastic} & 0.800 & 23.91 & 0.129 & 157.9 & 22.8M\\ 
			SAVP~\cite{lee2018stochastic} & 0.699 & 23.79 & 0.126 & 183.7 & 17.6M\\
			SAVP-VAE~\cite{lee2018stochastic} & 0.806 & 26.00 & 0.116 & 145.7 & 7.3M\\
			Struct-VRNN~\cite{minderer2019unsupervised} & 0.766 & 24.29 & 0.124 & 395.0 & 2.3M\\ 
			Grid keypoint (ours) & \textbf{0.837} & \textbf{27.11} & \textbf{0.092} & \textbf{144.2} & 2.0M\\
			\bottomrule
		\end{tabular}
	}
	\vspace{-4mm}
\end{table}

\begin{figure*}[t]
	\includegraphics[width=0.9\textwidth]{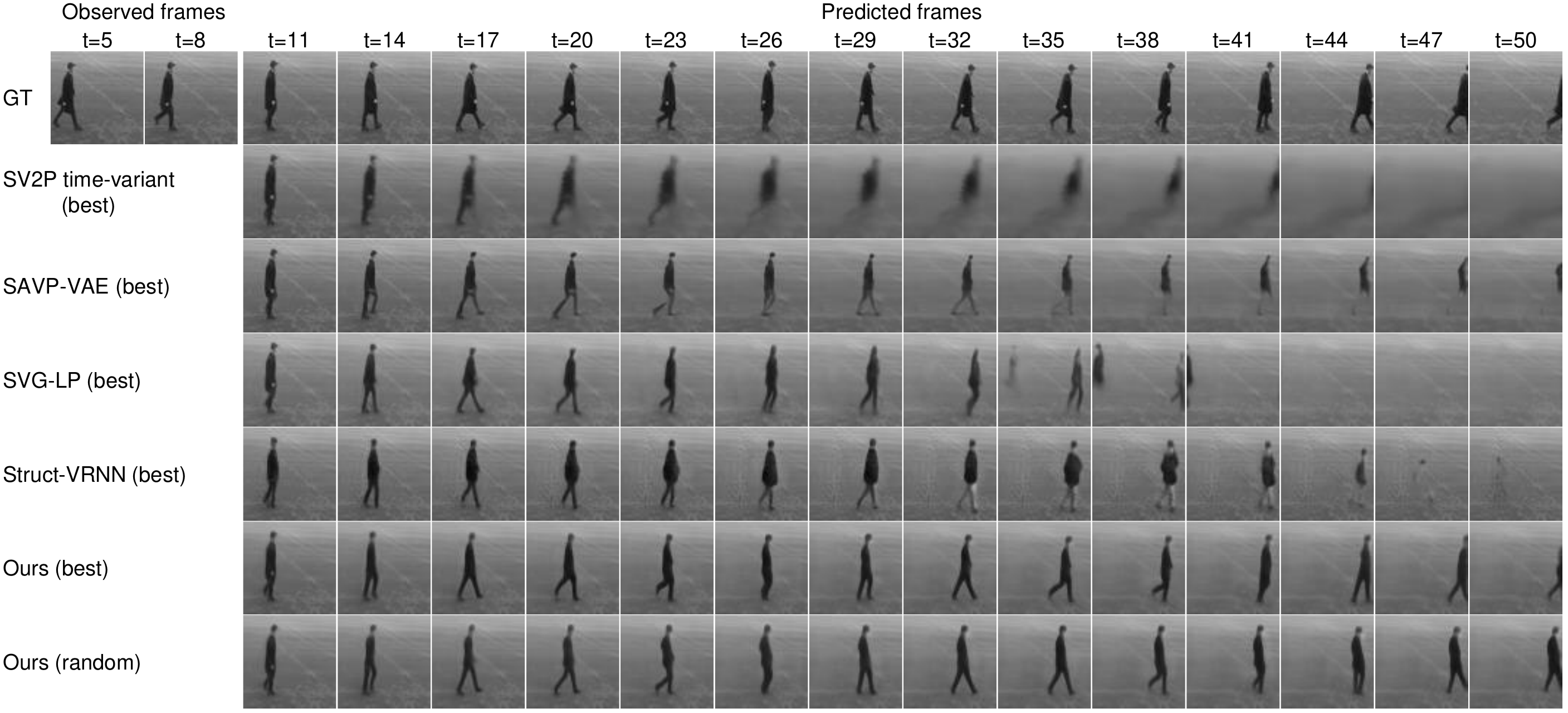}
	\centering
	\caption{Qualitative results on the KTH dataset. We show the best sample with the highest SSIM of different methods (best). We also present a random sample to demonstrate the diversity of our prediction model (random).
	}
	\label{fig:Qualitative_kth}
	\vspace{-4mm}
\end{figure*}

\subsubsection{Results on the KTH Dataset}
As shown in Table~\ref{tab:KTH} and Fig.~\ref{fig:kth_perframe}, our method significantly outperforms previous image-based stochastic video prediction methods on all frame-wise metrics. 
With large parameters, these methods give good results in short-term horizons, however, their performances deteriorate very quickly as time goes on because synthesis in the dense pixel-wise space tends to accumulate more errors.
Owing to our grid framework to diminish accumulated errors, our method achieves superior prediction quality and less deviation than the compared methods, especially in the long-term future. 
Additionally, our method attains performance boosts over the state-of-the-art keypoint-based method Struct-VRNN by a large margin.
Notably, our model also enjoys the least network parameter, which implies its promising prospect in large-scale applications.

We illustrate the qualitative results in Fig.~\ref{fig:Qualitative_kth}.
It is observed that image-based methods (SV2P, SAVP-VAE, and SVG-LP) tend to lose the person as time goes on, although SAVP-VAE gains an almost equal FVD score as ours.
The keypoint-based method Struct-VRNN also hardly preserves the person's shape in the long term and predicts gradually distorted frames due to the damage of keypoint spatial structures. 
Our model well preserves completeness and fidelity during a complete action period and can generate diverse and reasonable future frames (see the attached video for more examples).

\begin{figure}[t]
	\includegraphics[width=0.45\textwidth]{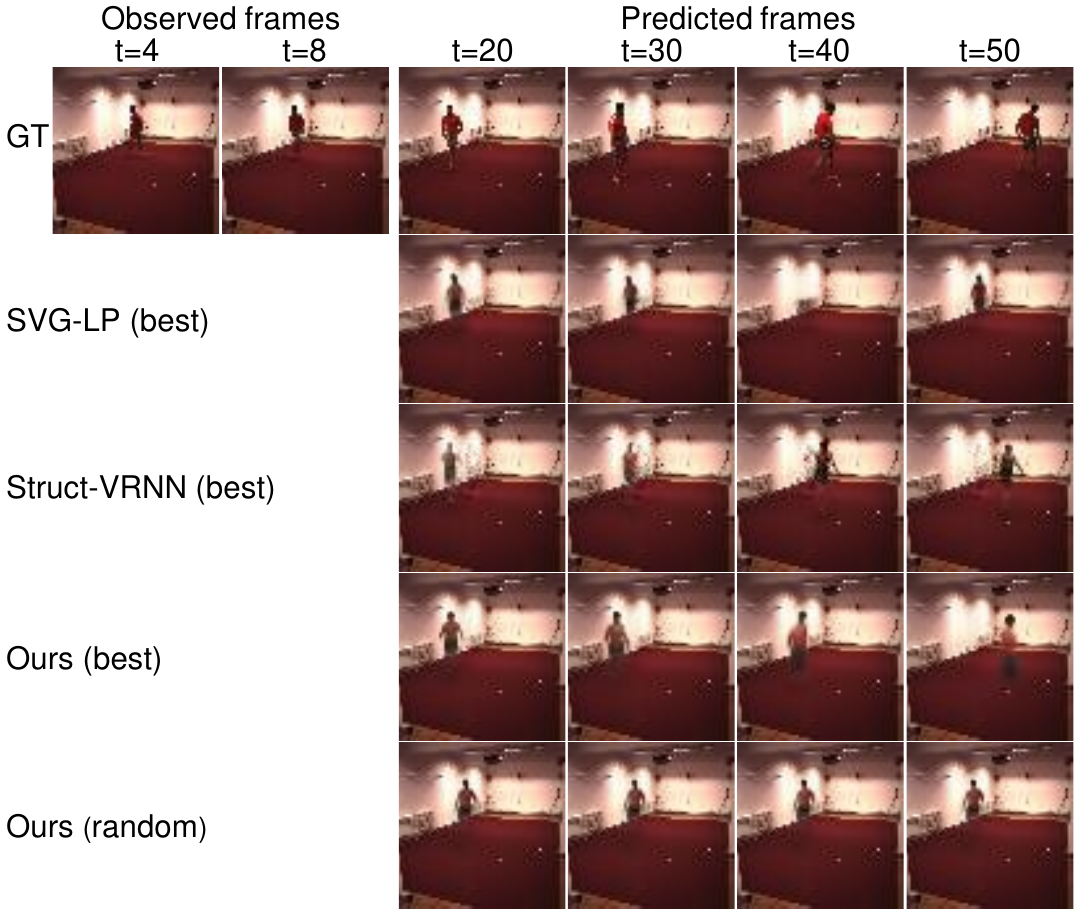}
	\centering
	\caption{Visual results on the Human3.6M dataset.  
	}
	\label{fig:human}
	\vspace{-4mm}
\end{figure}

\subsubsection{Results on the Human3.6M Dataset}
The results are reported in Table~\ref{tab:human} and note that we did not list the results of SV2P and SAVP given their pre-trained models on this dataset are not available.
We observe that SVG-LP gives inferior results to keypoint-based methods due to the difficulty of modeling long-term movements in pixel-level space.
Our method attains the best performance on all four metrics with the least model parameters.
The qualitative results are presented in Fig.~\ref{fig:human}.
We see that SVG-LP gives inconsistent predictions and even loses the person at the time step of 40.
Struct-VRNN fails to preserve the dress information of the person and generates frames with artifacts in the background.
As shown in both best and random samples, our model achieves consistently reasonable and various predictions.

\subsection{Effectiveness of Key Components}
\label{sec:ablation}
We progressively evaluate the effectiveness of our critical components in frame reconstruction ($\mathbf V'$) and future frame prediction ($\mathbf{\hat V}$) by answering the following questions: 
i) does our grid keypoint detection style improve the representation ability of keypoints? 
ii) does our grid keypoint prediction method boost the propagation accuracy of keypoint coordinates, thereby promoting video prediction performances?

\begin{table}[t]
	\centering
	\caption{Quantitative comparisons on the Human3.6M dataset. The best results under each metric are marked in bold.
	}
	\label{tab:human}
	\resizebox{0.4\textwidth}{!}{
		\begin{tabular}{l|ccccc}
			\toprule
			Method
			& SSIM$\uparrow$ & PSNR$\uparrow$ & LPIPS$\downarrow$ & FVD$\downarrow$ & \#param\\ 
			\midrule
			SVG-LP~\cite{denton2018stochastic} & 0.893 & 24.67 & 0.084 & 179.5 & 22.8M\\
			Struct-VRNN~\cite{minderer2019unsupervised} & 0.901 & 24.98 & 0.056 & 193.8 & 2.3M\\ 
			Grid keypoint (ours) & \textbf{0.915} & \textbf{26.06} & \textbf{0.055} & \textbf{166.1}& 2.0M\\ 
			\bottomrule
		\end{tabular}
	}
	\vspace{-4mm}
\end{table}

\subsubsection{Different Keypoint Detection Methods}
\label{sec:ablation_rec}
We first investigate the effectiveness of crucial components in keypoint detection by illustrating the performance of frame reconstruction.
We design the following ablation settings: 
i) baseline: only employing reconstruction loss $\mathcal{L}_\mathrm{rec}$ to detect keypoints in continuous space; 
ii) baseline + $\mathcal{L}_\mathrm{con}$: adding condensation loss $\mathcal{L}_\mathrm{con}$ to detect keypoints in continuous space; 
iii) baseline + gridding: only using $\mathcal{L}_\mathrm{rec}$ and detecting keypoints in finite grid space; 
iv) our full model: adding $\mathcal{L}_\mathrm{con}$ and detecting keypoints in grid space. 
We also include the detection part of Struct-VRNN~\cite{minderer2019unsupervised} for comparison, which employs a $(x,y,\mu)$-triplet to denote coordinate and scale.

\begin{figure}[t]
	\includegraphics[width=0.45\textwidth]{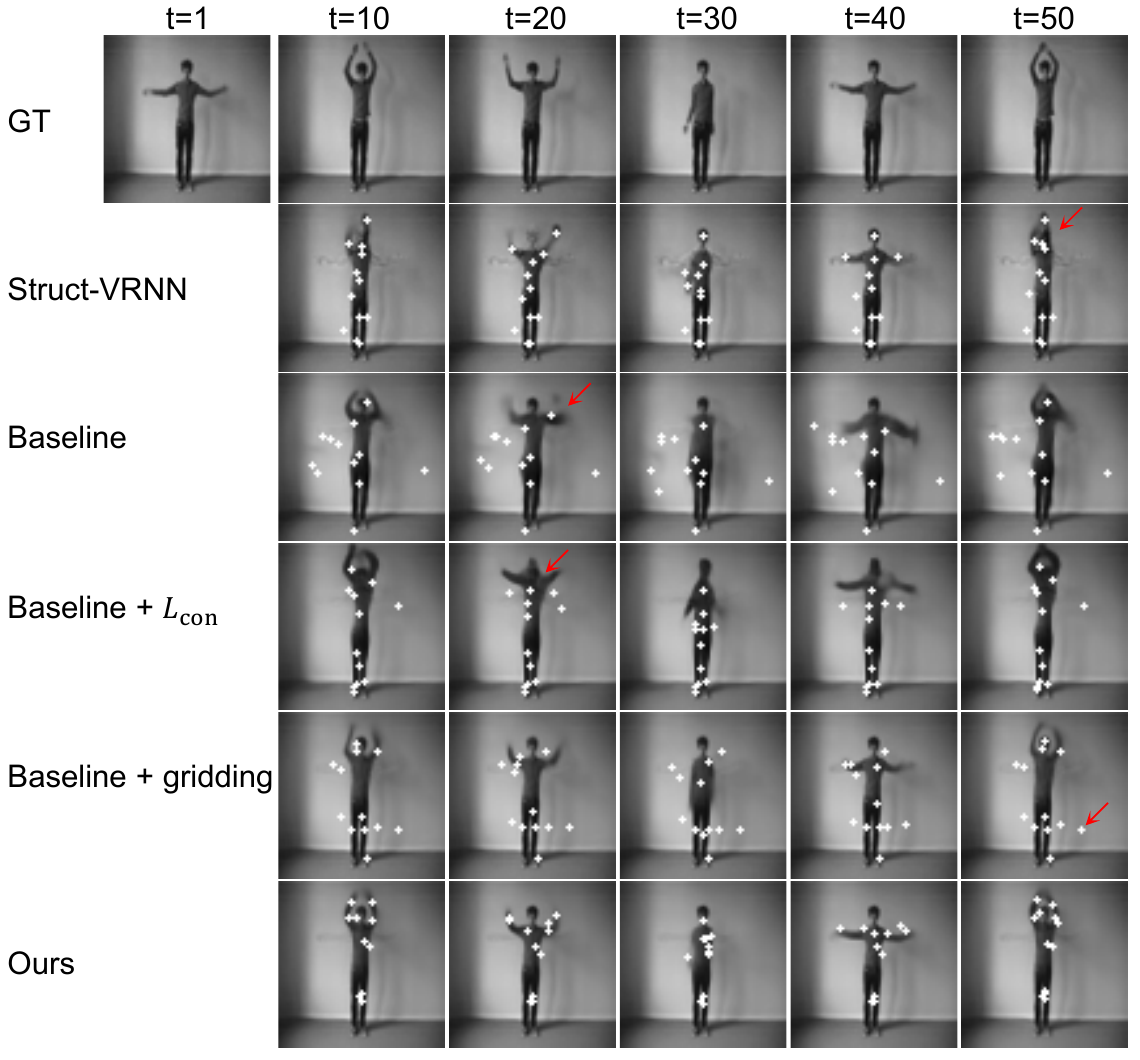}
	\centering
	\caption{Reconstruction results ($\mathbf V'$) of each method to indicate their representative capabilities on the KTH dataset. 
	}
	\label{fig:kth_keypoints_rec}
	\vspace{-1mm}
\end{figure}

The results on the KTH dataset are shown in Table~\ref{tab:ablation} and Fig.~\ref{fig:kth_keypoints_rec}.
We see that compared with baseline, either sub-modules contributes to improvements in keypoint representation for better frame restoration, boosting SSIM from 0.759 to 0.805 and 0.855, respectively.
As shown in Fig.~\ref{fig:kth_keypoints_rec}, $\mathcal{L}_\mathrm{con}$ encourages the model to concentrate the keypoints on the foreground region and bypasses the keypoint diffusion on the trivial background (see the third and fifth rows).
Our gridding regularization enhances the representation capability to reconstruct the more complete frames (see the second to fourth rows where the head or arms of the person tend to miss).
Equipped with both key components, our full model achieves the best keypoint representation, peaking SSIM at 0.862 with the reconstructed frames closest to ground truths.

\begin{table}[t]
	\centering
	\caption{Frame reconstruction results of different detection methods on the KTH dataset.}
	\label{tab:ablation}
	\resizebox{0.4\textwidth}{!}{
		\begin{tabular}{l|ccc}
\toprule
Method & SSIM$\uparrow$ & PSNR$\uparrow$ & LPIPS$\downarrow$ \\ \midrule
Struct-VRNN~\cite{minderer2019unsupervised} & 0.821          & 27.86          & 0.089    \\ 
Baseline                                             &      0.759     &    24.93       &    0.179          \\
Baseline + $\mathcal{L}_\mathrm{con}$                                            & 0.805          & 25.23          & 0.114             \\
Baseline + gridding                          &  0.855         & 29.31          &    0.095          \\
Baseline + $\mathcal{L}_\mathrm{con}$ + gridding (ours)                                                        & \textbf{0.862} & \textbf{29.68} & \textbf{0.076}             \\ \bottomrule
\end{tabular}
	}
	\vspace{-4mm}
\end{table}

\begin{figure}[t]
	\includegraphics[width=0.45\textwidth]{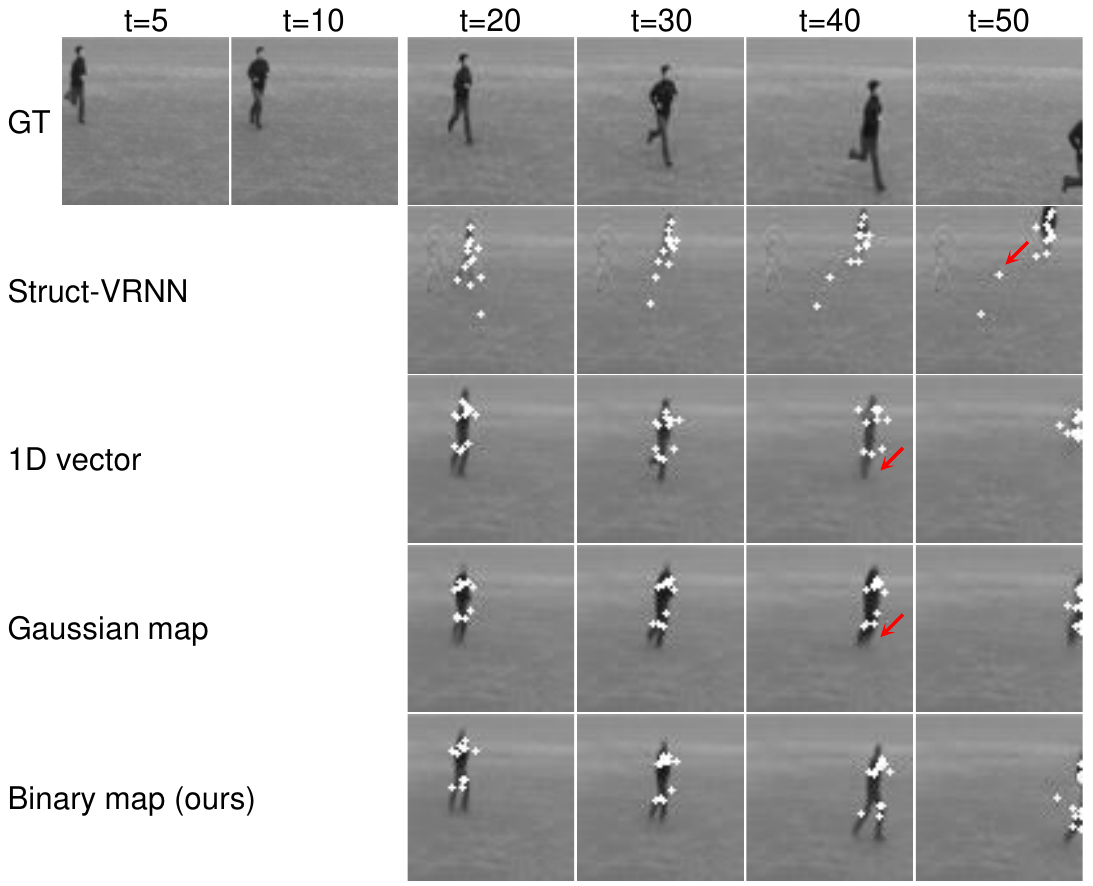}
	\centering
	\caption{Future frame generation results ($\mathbf{\hat V}$) from different keypoint prediction methods on the KTH dataset. 
	}
	\label{fig:kth_keypoints_pred}
	\vspace{-1mm}
\end{figure}

\begin{table}[t]
	\centering
	\caption{Image degeneration rate of different keypoint propagation methods on the KTH dataset.}
	\label{tab:ablation_degen}
	\resizebox{0.3\textwidth}{!}{
		\begin{tabular}{l|ccc}
\toprule
                                             Method     & SSIM             & PSNR              & LPIPS             \\ \midrule
Struct-VRNN~\cite{minderer2019unsupervised} & 6.7\%                           & 12.8\%                           & 39.8\%                           \\ 
1D vector                                                         & 5.2\%                           & 12.1\%                           & 46.4\%                           \\
Gaussian map                                                      &   5.7\%                             &     12.9\%                            &     50.1\%                            \\
Binary map (ours)                                                 & \textbf{2.9\%} & \textbf{8.7\%} & \textbf{21.0\%} \\ \bottomrule
\end{tabular}
	}
	\vspace{-2mm}
\end{table}

\begin{table}[th!]
	\centering
	\caption{Average coordinate prediction error in the grid space with 95\% confidence interval on the KTH dataset.
	}
	\label{tab:coordinate_error}
	\resizebox{0.45\textwidth}{!}{
		\begin{tabular}{l|cccc}
\toprule
   Method & $t=$ 20       & $t=$ 30       & $t=$ 40       & $t=$ 50       \\ \midrule
Struct-VRNN~\cite{minderer2019unsupervised} & 4.75$\pm$0.13 & 5.39$\pm$0.22 & 6.07$\pm$0.29 & 8.24$\pm$0.52 \\
1D vector             & 2.87$\pm$0.14 & 3.36$\pm$0.21 & 3.94$\pm$0.34 & 5.49$\pm$0.60 \\
Gaussian map             & 3.01$\pm$0.15 & 3.89$\pm$0.29 & 4.57$\pm$0.45 & 5.99$\pm$0.78 \\
Binary map (ours)             & \textbf{2.43$\pm$0.18} & \textbf{3.07$\pm$0.33} & \textbf{3.49$\pm$0.38} & \textbf{4.60$\pm$0.55} \\ \bottomrule
\end{tabular}
	}
	\vspace{-4mm}
\end{table}

\subsubsection{Different Keypoint Propagation Styles}
\label{sec:ablation_pred}
We then investigate how our gridding regularization helps retain keypoint structures during coordinate propagation for future frame prediction. 
To purely validate the effectiveness for propagation, we design the following configurations that are all based on the best-detected keypoints from our full detection model, and we only vary the keypoint representation styles in propagation:
i) 1D vector: directly using a 1D vector to represent keypoint coordinates and an LSTM to model dynamics;
ii) Gaussian map: transforming keypoint coordinates to Gaussian maps \gao{(the form used in keypoint detection)} and using a convLSTM to model dynamics; 
iii) Binary map: changing to our proposed binary maps and using a convLSTM. 
We also compare with Struct-VRNN, with both its detection and prediction parts unchanged. 

Video prediction performances of keypoint-based methods are upper-bounded by their reconstruction quality.
We first show the deterioration rate of predicted frames relative to the corresponding upper-bounds of different settings on the KTH dataset.
We see that from Table~\ref{tab:ablation_degen}, our binary map with convLSTM achieves the least performance degradation. Though employing convLSTM, representing keypoint as Gaussian map shall reversely affect the propagation given much uncertain and ambiguous information involved in this style.
This observation demonstrates that our gridding regularization is indispensable to yield the efficacy of convLSTM to preserve the spatial structure.

We further provide a more intuitive and comprehensive analysis by calculating the coordinate errors at different prediction time steps.
The error is measured by grid distances averaged over each keypoint between predicted keypoint coordinates and their ground truth positions, i.e., keypoints produced by our detection model. 
The results are reported in Table~\ref{tab:coordinate_error}.
We see that the prediction error in all three settings grows slower than Struct-VRNN, demonstrating that our method provides a more substantial representation base in the detected keypoints for further propagation.
Our proposed binary map further beats other settings in all time steps with apparent gaps.  
We also illustrate the qualitative results in Fig.~\ref{fig:kth_keypoints_pred}.
We observe that our method can sufficiently hold the complete information in the long-range prediction.

\begin{table}[t]
	\centering
	\caption{Ablative testing results for different number of keypoints.}
	\label{tab:ablation_num_kp}
	\resizebox{0.45\textwidth}{!}{
		\begin{tabular}{c|ccc|ccc}
\toprule
\multirow{2}{*}{Keypoint number} & \multicolumn{3}{c|}{Reconstruction ($\mathbf V'$)}            & \multicolumn{3}{c}{Prediction ($\mathbf{\hat V}$)}               \\ \cline{2-7} 
                                 & SSIM$\uparrow$ & PSNR$\uparrow$ & LPIPS$\downarrow$ & SSIM$\uparrow$ & PSNR$\uparrow$ & LPIPS$\downarrow$ \\ \midrule
6                                & 0.848          & 28.76          & 0.105             & 0.833          & 27.06          & 0.110             \\
12                               & \textbf{0.862}          & \textbf{29.68}          & \textbf{0.076}             & \textbf{0.837}          & \textbf{27.11}          & \textbf{0.092}             \\
18                               & 0.854          & 29.42          & 0.094             & 0.819          & 26.42          & 0.113             \\ \bottomrule
\end{tabular}
	}
	\vspace{-1mm}
\end{table}

\subsubsection{Different Numbers of Keypoints}
\label{sec:ablation_num}
We also analyze the impact of different keypoint numbers for frame reconstruction and prediction on the KTH dataset.
The results using 6, 12, and 18 keypoints are listed in Table~\ref{tab:ablation_num_kp}.
We see that slight performance improvements are gained when increasing the keypoints from 6 to 12.
However, results decrease especially for SSIM after the keypoint number further increasing to 18. 
\gao{The reason might be that deficient keypoints could not represent the key video dynamics, and excessive keypoints lead to overfitting on trivial details. 
Experiments on other datasets also verify this observation.}
Therefore, we choose 12 keypoints to implement our method.

\subsection{Results on Robot-assisted Surgical Videos}
\begin{figure}[t]
	\includegraphics[width=0.45\textwidth]{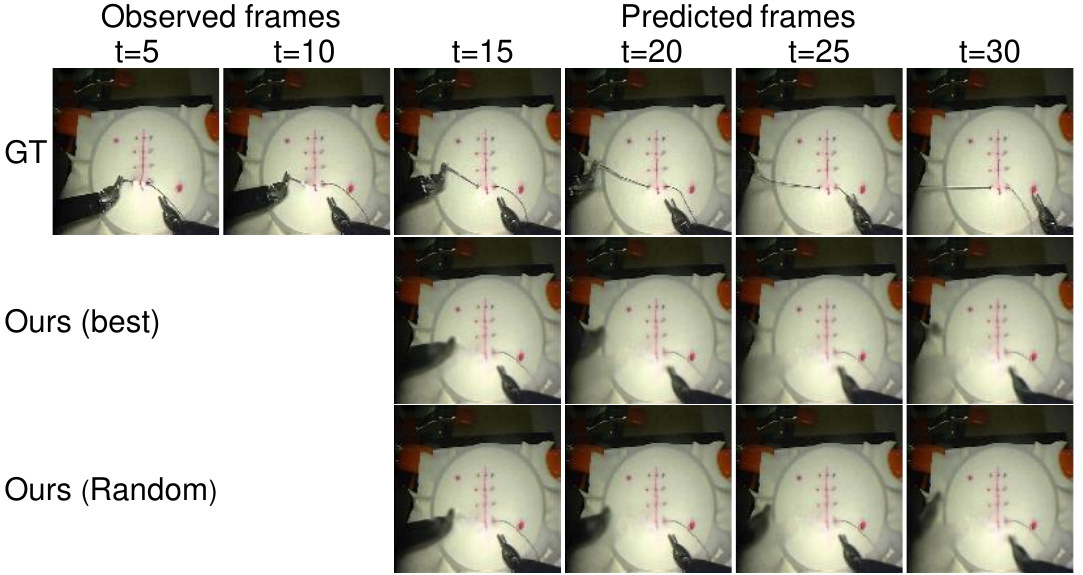}
	\centering
	\caption{Qualitative results on the JIGSAWS dataset.  
	}
	\label{fig:jigsaws}
	\vspace{-4mm}
\end{figure}

Our keypoint-based method also enjoys significant advantages to deploy in robots, given its lightweight model scale.
We evaluate our method on the JIGSAWS~\cite{gao2014jhu}, a complicated real-world robot-assisted surgery dataset.
It contains surgical robotic motions on the dual-arm \textit{da Vinci} robot system~\cite{Freschi2013}.
As shown in Fig.~\ref{fig:jigsaws}, our method generates promising qualitative results with diverse and reasonable movements of robotic arms, demonstrating the great potential of our approach for robotic applications.

\section{Conclusion}

In this paper, we propose a new grid keypoint learning framework for stochastic video prediction. 
We detect discrete keypoints in a grid space, which are further regularized by our condensation loss to encourage explainable high-level configurations. 
Owing to our proposed binary maps, accurate keypoint coordinate prediction in a long-term horizon is realized to improve the transformed future frames. 
We validate our approach on several popular datasets and show the superior results of our method with high parameter-efficiency in terms of both quantitative and qualitative evaluations.
For future works, we plan to investigate the potential of our framework to deal with skeleton data and explore its promising applications for humans or robots.



\bibliographystyle{IEEEtran}
\bibliography{ref}

\begin{thebibliography}{10}
\providecommand{\url}[1]{#1}
\csname url@rmstyle\endcsname
\providecommand{\newblock}{\relax}
\providecommand{\bibinfo}[2]{#2}
\providecommand\BIBentrySTDinterwordspacing{\spaceskip=0pt\relax}
\providecommand\BIBentryALTinterwordstretchfactor{4}
\providecommand\BIBentryALTinterwordspacing{\spaceskip=\fontdimen2\font plus
\BIBentryALTinterwordstretchfactor\fontdimen3\font minus
  \fontdimen4\font\relax}
\providecommand\BIBforeignlanguage[2]{{%
\expandafter\ifx\csname l@#1\endcsname\relax
\typeout{** WARNING: IEEEtran.bst: No hyphenation pattern has been}%
\typeout{** loaded for the language `#1'. Using the pattern for}%
\typeout{** the default language instead.}%
\else
\language=\csname l@#1\endcsname
\fi
#2}}

\bibitem{byeon2018contextvp}
W.~Byeon \emph{et~al.}, ``{ContextVP: Fully context-aware video prediction},''
  in \emph{ECCV}, 2018.

\bibitem{kumar2020videoflow}
M.~Kumar \emph{et~al.}, ``Videoflow: A conditional flow-based model for
  stochastic video generation,'' in \emph{ICLR}, 2020.

\bibitem{finn2017deep}
C.~Finn \emph{et~al.}, ``Deep visual foresight for planning robot motion,'' in
  \emph{ICRA}, 2017.

\bibitem{jin2018varnet}
B.~Jin \emph{et~al.}, ``{VarNet: Exploring variations for unsupervised video
  prediction},'' in \emph{IROS}, 2018.

\bibitem{gao2020automatic}
X.~Gao \emph{et~al.}, ``Automatic gesture recognition in robot-assisted surgery
  with reinforcement learning and tree search,'' in \emph{ICRA}, 2020.

\bibitem{villegas2017learning}
R.~Villegas \emph{et~al.}, ``Learning to generate long-term future via
  hierarchical prediction,'' in \emph{ICML}, 2017.

\bibitem{jin2020exploring}
B.~Jin \emph{et~al.}, ``Exploring spatial-temporal multi-frequency analysis for
  high-fidelity and temporal-consistency video prediction,'' in \emph{CVPR},
  2020.

\bibitem{castrejon2019improved}
L.~Castrejon \emph{et~al.}, ``{Improved conditional VRNNs for video
  prediction},'' in \emph{ICCV}, 2019.

\bibitem{villegas2019high}
R.~Villegas \emph{et~al.}, ``High fidelity video prediction with large
  stochastic recurrent neural networks,'' in \emph{NeurIPS}, 2019.

\bibitem{jakab2018unsupervised}
T.~Jakab \emph{et~al.}, ``Unsupervised learning of object landmarks through
  conditional image generation,'' in \emph{NeurIPS}, 2018.

\bibitem{zhang2018unsupervised}
Y.~Zhang \emph{et~al.}, ``Unsupervised discovery of object landmarks as
  structural representations,'' in \emph{CVPR}, 2018.

\bibitem{minderer2019unsupervised}
M.~Minderer \emph{et~al.}, ``Unsupervised learning of object structure and
  dynamics from videos,'' in \emph{NeurIPS}, 2019.

\bibitem{kalchbrenner2017video}
N.~Kalchbrenner \emph{et~al.}, ``Video pixel networks,'' in \emph{ICML}, 2017.

\bibitem{tulyakov2018mocogan}
S.~Tulyakov \emph{et~al.}, ``{MoCoGAN: Decomposing motion and content for video
  generation},'' in \emph{CVPR}, 2018.

\bibitem{chung2015recurrent}
J.~Chung \emph{et~al.}, ``A recurrent latent variable model for sequential
  data,'' in \emph{NeurIPS}, 2015.

\bibitem{babaeizadeh2018stochastic}
M.~Babaeizadeh \emph{et~al.}, ``Stochastic variational video prediction,'' in
  \emph{ICLR}, 2018.

\bibitem{denton2018stochastic}
E.~Denton \emph{et~al.}, ``Stochastic video generation with a learned prior,''
  in \emph{ICML}, 2018.

\bibitem{lee2018stochastic}
A.~X. Lee \emph{et~al.}, ``Stochastic adversarial video prediction,''
  \emph{arXiv preprint arXiv:1804.01523}, 2018.

\bibitem{franceschi2020stochastic}
J.-Y. Franceschi \emph{et~al.}, ``Stochastic latent residual video
  prediction,'' in \emph{ICML}, 2020.

\bibitem{kim2019unsupervised}
Y.~Kim \emph{et~al.}, ``Unsupervised keypoint learning for guiding
  class-conditional video prediction,'' in \emph{NeurIPS}, 2019.

\bibitem{silver2017mastering}
D.~Silver \emph{et~al.}, ``{Mastering the game of Go without human
  knowledge},'' \emph{Nature}, 2017.

\bibitem{shi2015convolutional}
X.~Shi \emph{et~al.}, ``{Convolutional LSTM network: A machine learning
  approach for precipitation nowcasting},'' in \emph{NeurIPS}, 2015.

\bibitem{kingma2013auto}
D.~P. Kingma \emph{et~al.}, ``Auto-encoding variational bayes,'' in
  \emph{ICLR}, 2014.

\bibitem{schuldt2004recognizing}
C.~Schuldt \emph{et~al.}, ``{Recognizing human actions: a local SVM
  approach},'' in \emph{ICPR}, 2004.

\bibitem{ionescu2013human3}
C.~Ionescu \emph{et~al.}, ``{Human3.6M: Large scale datasets and predictive
  methods for 3D human sensing in natural environments},'' \emph{IEEE Trans.
  Pattern Anal. Machine Intell.}, 2013.

\bibitem{wang2004image}
Z.~Wang \emph{et~al.}, ``Image quality assessment: from error visibility to
  structural similarity,'' \emph{IEEE Trans. Image Processing}, 2004.

\bibitem{zhang2018unreasonable}
R.~Zhang \emph{et~al.}, ``The unreasonable effectiveness of deep features as a
  perceptual metric,'' in \emph{CVPR}, 2018.

\bibitem{unterthiner2018towards}
T.~Unterthiner \emph{et~al.}, ``{Towards accurate generative models of video: A
  new metric \& challenges},'' \emph{arXiv preprint arXiv:1812.01717}, 2018.

\bibitem{kingma2014adam}
D.~P. Kingma \emph{et~al.}, ``{Adam: A method for stochastic optimization},''
  in \emph{ICML}, 2014.

\bibitem{gao2014jhu}
Y.~Gao \emph{et~al.}, ``{JHU-ISI gesture and skill assessment working set
  (JIGSAWS): A surgical activity dataset for human motion modeling},'' in
  \emph{MICCAI Workshop}, 2014.

\bibitem{Freschi2013}
C.~Freschi \emph{et~al.}, ``{Technical review of the da Vinci surgical
  telemanipulator},'' \emph{Int J Med Robot}, 2013.

\end{thebibliography}

\end{document}